\documentclass{article}
\usepackage{arxiv}

\usepackage{graphicx}
\usepackage[utf8]{inputenc} 
\usepackage[T1]{fontenc}    
\usepackage{hyperref}       
\usepackage{url}            
\usepackage{booktabs}       
\usepackage{amsfonts}       
\usepackage{nicefrac}       
\usepackage{microtype}      
\usepackage{lipsum}
\usepackage{subcaption}
\usepackage{multirow}

\title{Object Detection on Single Monocular Images through Canonical Correlation Analysis}

\author{
  Zifan Yu\\
  University of Southern California\\
  Los Angeles, California\\
  \texttt{zifanyu@usc.edu} \\
   \And
 Suya You \\
  U.S. Army Research Laboratory\\
  Playa Vista, California\\
  \texttt{suya.you.civ@mail.mil} \\
}

\begin{document}
\maketitle

\begin{abstract}
Without using extra 3-D data like points cloud or depth images for providing 3-D information, we retrieve the 3-D object information from single monocular images.  The high-quality predicted depth images are recovered from single monocular images, and it is fed into the 2-D object proposal network with corresponding monocular images. Most existing deep learning frameworks with two-streams input data always fuse separate data by concatenating or adding, which views every part of a feature map can contribute equally to the whole task. However, when data are noisy, and too much information is redundant, these methods no longer produce predictions or classifications efficiently. In this report, we propose a two-dimensional CCA(canonical correlation analysis) framework to fuse monocular images and corresponding predicted depth images for basic computer vision tasks like image classification and object detection. Firstly, we implemented different structures with one-dimensional CCA and Alexnet to test the performance on the image classification task. And then, we applied one of these structures with 2D-CCA for object detection. During these experiments, we found that our proposed framework behaves better when taking predicted depth images as inputs with the model trained from ground truth depth.
\end{abstract}


\section{Introduction}
    With the help of state-of-art deep learning baselines, especially the combination of Faster R-CNN and Features Pyramid Network(FPN), many novel methods achieved higher accuracy for 2-D object detection \cite{Mask RCNN,Faster RCNN,FPN}. For the 3-D object detection, except for an efficient 2-D object backbone, low-cost and reliable depth information are needed to generate considerable predictions. However, in most of the cases, such depth information is captured by expensive and sophisticated hardware. Recently, because of the improvements achieved in terms of recovery depth information from monocular images or stereo images, accurate 3D object detection only with monocular images becomes possible \cite{TLNet, DDepth, SDepth, EDepth}. Fusion two-streams information efficiently becomes the next focus. Compared to previous fusion methods, we used the Canonical Correlation Analysis(CCA) to extract the projected canonical feature maps instead of directly concatenating feature maps of two views \cite{CCA} . We used a dimension-based custom layer to approximate the effect of two-dimensional CCA \cite{2DCCA}, and this structure was first proposed by \cite{CCAR}. \cite{CCAR} utilized this structure on image classification tasks with one-dimensional CCA. Therefore, our structure fits other computer vision tasks that need 2-D feature maps for classifiers or regressors. 
    
    For experiments, we tested our structure on Virtual KITTI \cite{vkitti}, KITTI\cite{ kitti_object,kitti}, and CityScape \cite{cscape}. KITTI and Cityscape are used for visually quality analysis. We used three evaluation metrics for quantitative analysis with Virtual Kitti, they are mAP, mRecall, and mIoU, respectively, and they will be introduced in detail in the experiments section. For the depth predictions part, we utilized the network structure and pre-trained model on KITTI of \cite{DDepth}, which only need monocular images as input. Therefore, in our structure, depth estimation is an external module without modifying. 
    
    Figure 1 shows the whole pipeline we designed for 3D object detection, and this paper focuses on feature extraction and 2-D object proposal part. Figure 2 shows a ground truth depth image from Virtual KITTI, and its corresponding predicted depth image with DenseDepth \cite{DDepth}. It is evident that the predicted depth image expresses essential information like cars.  The last two images are the results of 2-D proposals of each method.  

\begin{figure}[ht]
  \centering
    \includegraphics[width=0.75\paperwidth]{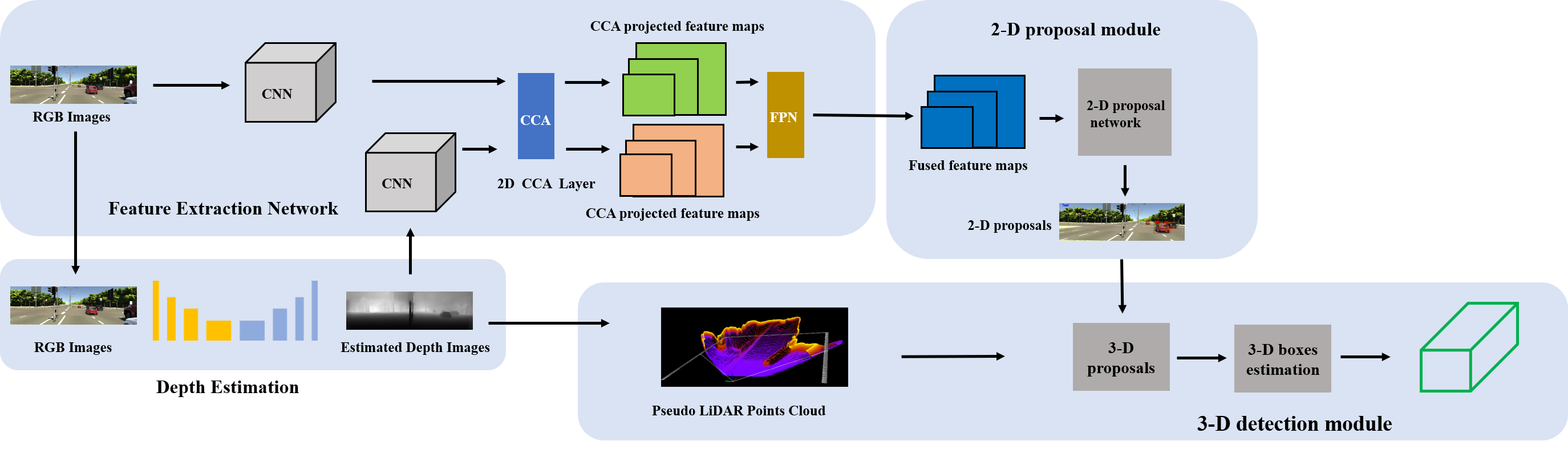}
  \caption{Pipeline for 3-D object detection \small This report focus on the feature extraction network and the 2-D proposal module.}
  \label{fig:network}
\end{figure}

\begin{figure}[h!]
  \begin{subfigure}[b]{0.5\linewidth}
    \centering
    \includegraphics[width=0.5\linewidth]{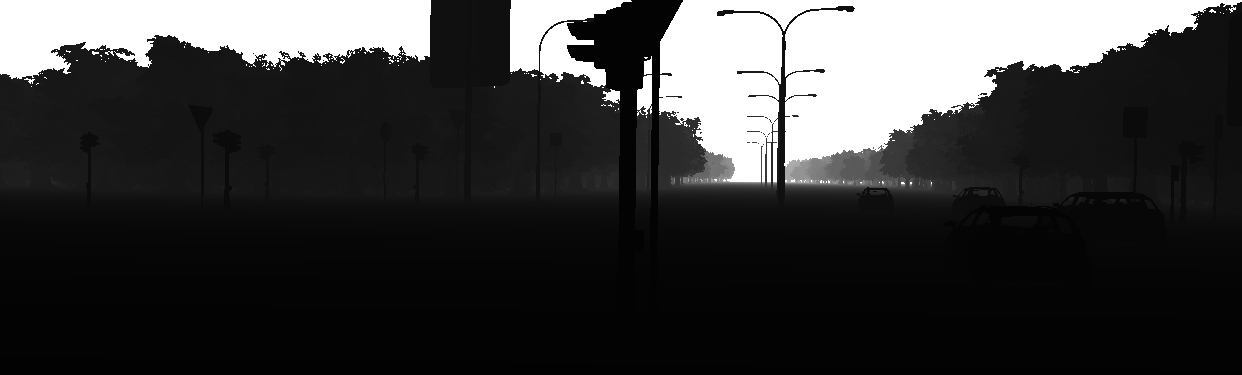}
    \caption{Ground Truth Depth}
  \end{subfigure}
  \begin{subfigure}[b]{0.5\linewidth}
  \centering
    \includegraphics[width=0.5\linewidth]{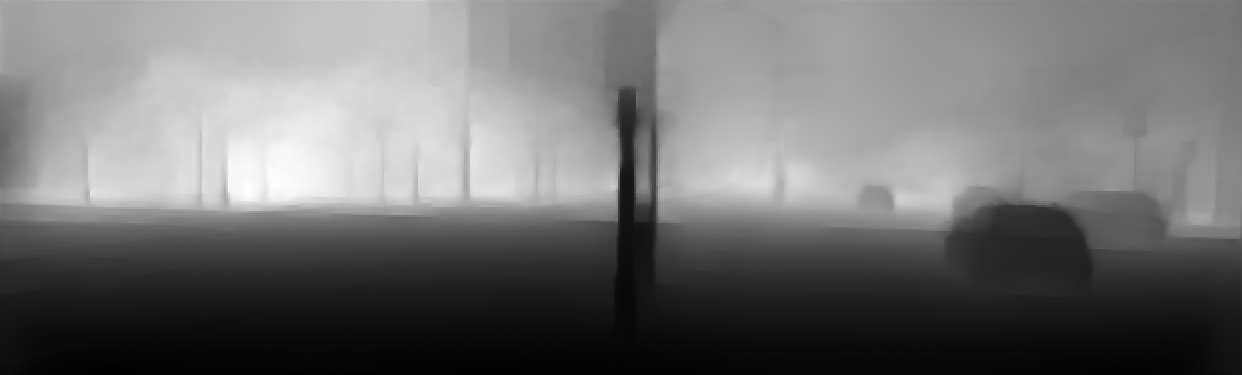}
    \caption{Dense Depth}
  \end{subfigure}
  \begin{subfigure}[b]{0.5\linewidth}
  \centering
    \includegraphics[width=0.5\linewidth]{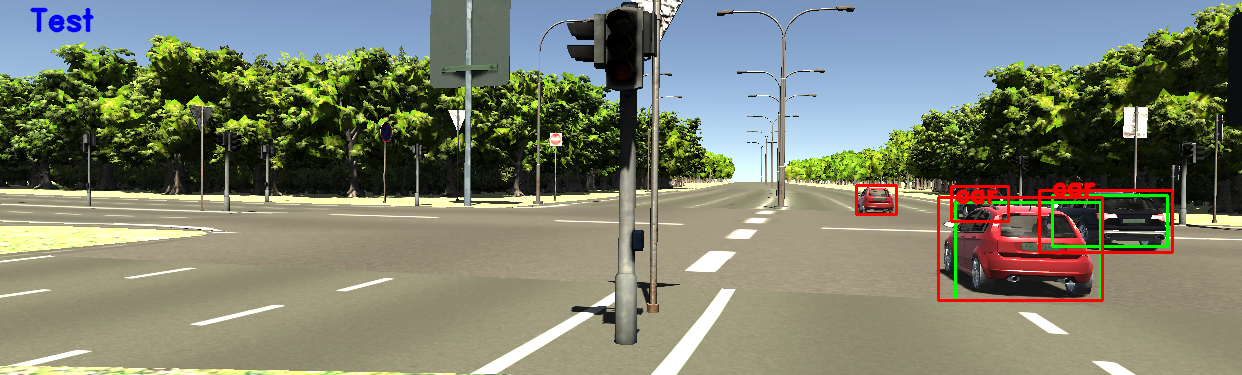}
    \caption{Predictions of baseline model}
  \end{subfigure}
  \begin{subfigure}[b]{0.5\linewidth}
  \centering
    \includegraphics[width=0.5\linewidth]{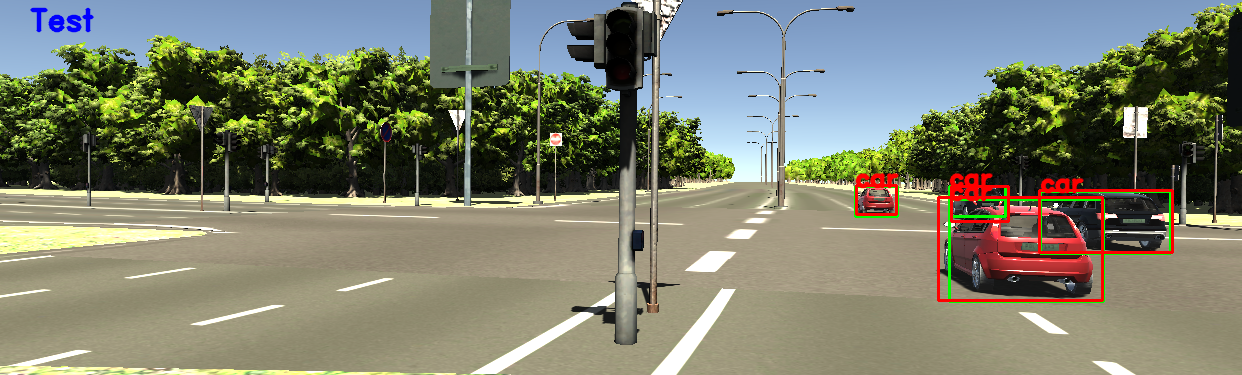}
    \caption{Predictions of model with CCA}
  \end{subfigure}
  \caption{\small The first image is the ground truth depth image, and the second one is the corresponding depth image predicted by Dense Depth. The last two images show the difference between models with CCA and baseline method when taking predicted depth images as inputs.}
  \label{fig:image_group_1}
\end{figure}

\section{Related Works}
\label{sec:headings}

\subsection{Canonical Correlation Analysis}
CCA is a widely used tool to analyze the correlation relationships of two separate sets of features or variables \cite{CCA, CCAR}. Since it was first proposed by \cite{CCA}, many variants have been developed for non-linear calculation and its application with neural networks \cite{DCCA}. \cite{DCCA} first proposed a deep learning version of CCA, which uses the correlation of two sets of CCA projected variables as the objective function to minimize. As neural networks use small-batch based methods for optimization, the singularity problem of covariance matrices happens because of the small sample size compared to the dimensions of data space \cite{DCCA, CCA_Depth}. Most existing frameworks use similar ways with DCCA to calculate CCA on a relatively big batch to avoid the above problem \cite{CCAR, E2ECCA}. 

\cite{CCAR} uses a fully connected layer to approximate the calculation of CCA and avoid the above singularity problem on a small sample size by calculating canonical variables outside the neural network.  We replicate the image classification experiments that \cite{CCAR} did in our works, and we extend this method with two-dimensional CCA into the 2-D object detection by designing a dimension-based custom layer \cite{2DCCA}, detailed in Section 3.2.  

\subsection{2-D Object Detection}

There are two types of verified efficient baselines for 2-D object detection \cite{Detection_S}. One is named the two-stage methods like Fast R-CNN \cite{Fast RCNN}, Faster R-CNN \cite{Faster RCNN}, and Mask R-CNN \cite{Mask RCNN} with region proposals network(RPN). Another one is one-stage methods like YOLO \cite{YOLO}, RetinaNet \cite{Focal}, and Single-shot detector (SSD) \cite{SSD} without RPN. The feature pyramid network (FPN) enhances the detection accuracy of the above methods by efficient detecting objects with multi-dimensions \cite{FPN}. 
 
 Recent improvements of 2-D object detection focus on details of each sub-module in the above methods, like \cite{LRCNN} tries to balance the information of feature maps in different levels and the number of anchor boxes with different difficulty, and \cite{Anchor-Free} optimize the assignment of  anchor boxes for regressor, and \cite{relationnetwork} utilizes attention model to extract relations between each instance to improve classification accuracy. For our work, we use the depth information generated for 3-D boxes estimation to provide extract information for 2-D region proposals. 

\section{Methods}

Two main parts that our work cover - views fusion with multi-structures with one-dimensional CCA for image classification, and views fusion with two-dimensional CCA for 2-D region proposals. For the image classification task, we implemented multi-structures what was proposed by \cite{CCAR, E2ECCA}, and for 2-D object detection, we extend one of these methods into 2-D feature space and test the performance with estimated depth information. The following sections cover the details. 
\subsection{Views Fusion for Image Classification}

AlexNet \cite{alexnet} is the baseline method for feature extraction and one fully connected layer following the feature extraction module for each view.  Since the canonical variables have a lower dimension than the original feature variables, the fully connected layer projects the feature vector into a lower dimension space for comparison with the model that uses CCA to project the high dimension feature vector into a lower dimension. And then, the two projected feature vectors are concatenated for a classifier, which includes two fully connected layers. Figure 3 shows the whole structure of the baseline method for image classification. 

\begin{figure}[ht]
  \centering
    \includegraphics[width=0.6\paperwidth]{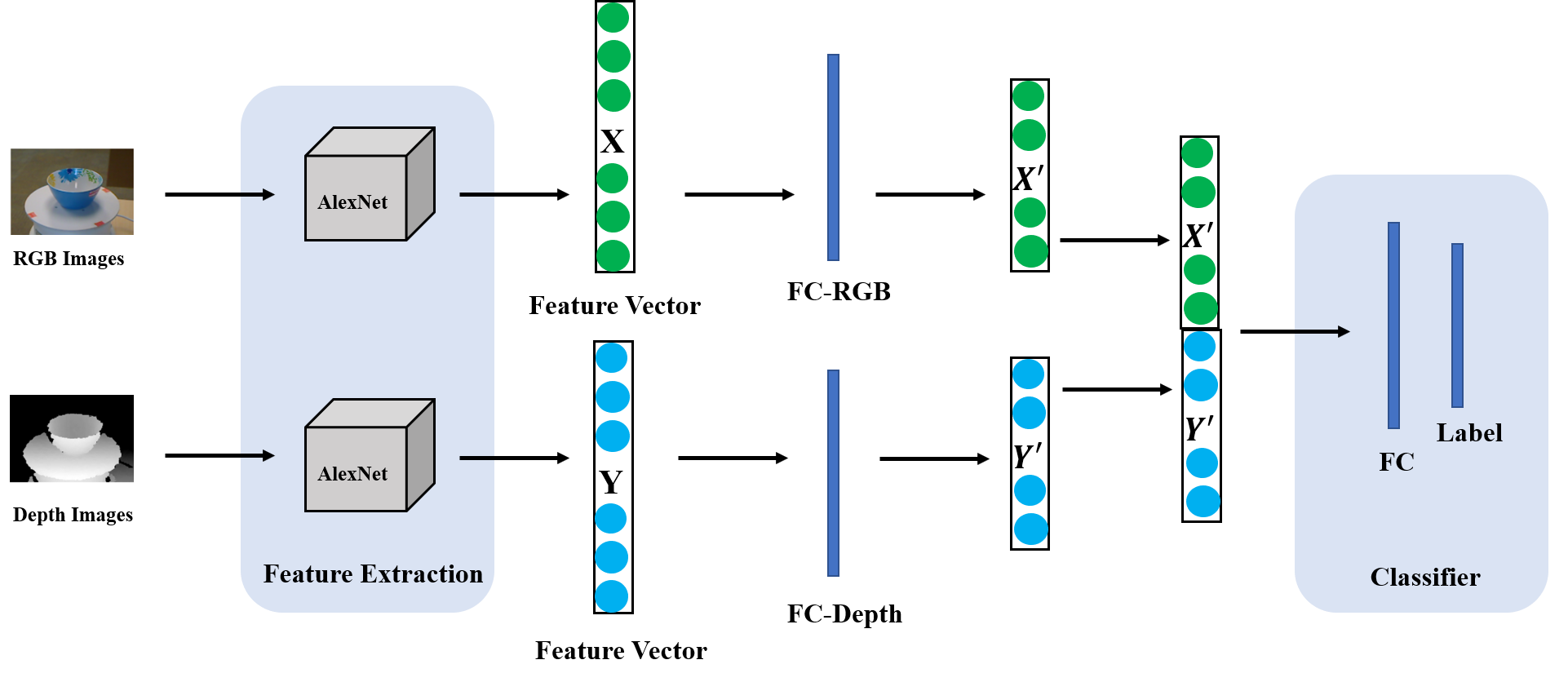}
  \caption{Baseline Method for Image Classification \small FC means a fully connected layer. The classifier is consists of two fully connected layers}
  \label{fig:network}
\end{figure}

The objective function of this method is the categorical cross-entropy loss. The output dimension of the last fully connected layer inside the classifier is equal to the number of classes. $L_{CCE}$ stands for the categorical cross-entropy loss.  

\subsubsection{Method 1: CCA Regularization}

Deep CCA  implemented a non-linear CCA by using the correlation of features from two views as a sub loss function. CCA Regularization method proposed by \cite{CCAR} also used the correlation of projected feature vectors as part of the objective function by integrating it into the categorical original cross-entropy loss as a regularization. In this way, the correlation gets maximized during training. For this method, the whole objective function can be expressed by equation (1) and (2). 

\begin{equation}
    L = L_{CCE} + \lambda L_{corr}
\end{equation}
\begin{equation}
    L_{corr} = -corr(U^TX, V^TY) = -corr(X^{\prime}, Y^{\prime})= - \frac{U^T\Sigma_{xy}V}{\sqrt{U^T\Sigma_{xx}UV^T\Sigma_{yy}V}}
\end{equation}

Where $X$ and $Y$ are two sets of variables, and the $U^TX$ and $V^TY$ are canonical variables, and $\Sigma_{xx}$ and $\Sigma_{yy}$ are covariance matrices for X and Y separately. The CCA aims to find the optimal pair of transformation so that the correlation between the two projected feature vectors is maximum. Therefore, the optimal pair of transformation can be expressed as:

\begin{equation}
    (U^*, V^*) = argmax_{U,V} \frac{U^T\Sigma_{xy}V}{\sqrt{U^T\Sigma_{xx}UV^T\Sigma_{yy}V}}
\end{equation}

Based on derivation presented in \cite{CCAR}, the optimal pair of transformation is equal to $(\Sigma_{xx}^{\frac{1}{2}}\tilde{U}_L, \Sigma_{yy}^{\frac{1}{2}}\tilde{V}_L) $, and $(\tilde{U}_L, \tilde{V}_L)$ is the first left and right singular vectors of matrix $T$, which $T$ is equal to $\Sigma_{xx}^{-(\frac{1}{2})}\Sigma_{xy}\Sigma_{yy}^{-(\frac{1}{2})}$. Furthermore, the gradient for network backpropagation over feature $X ^ \prime$ or $Y ^ \prime$ is stated in the follows:
\begin{equation}
\frac{\partial corr(X ^ \prime, Y ^ \prime)}{\partial X } = \frac{1}{N - 1}(2 \Delta_{11} \tilde{X } + \Delta_{12}\tilde{Y })
\end{equation}
\begin{equation}
\Delta_{11} = -\frac{1}{2}\Sigma_{11}^{-\frac{1}{2}}\tilde{U}D\tilde{U}^T\Sigma_{11}^{-\frac{1}{2}}
\end{equation}
\begin{equation}
\Delta_{12} = -\frac{1}{2}\Sigma_{11}^{-\frac{1}{2}}\tilde{U}\tilde{V}^T\Sigma_{22}^{-\frac{1}{2}}
\end{equation}
The gradient with respect to $Y$ has a similar form with the above one. As this method use the correlation as part of the total loss, the singularity problem over small-batch size mentioned above still exists, which means relatively big batch size is required to avoid this problem, and the backpropagation over correlation needs high-cost computation. 

\subsubsection{Method 2: CCA Regularization via Cost Amortization}

\cite{CCAR} proposed an advanced but low computation cost and a more reasonable structure to approximate the computation of CCA, which is named ACCAR. Compared to the first method, no backpropagation over complex matrix operations like SVD or singular value decomposition required in this structure, and the CCA pairs of transformation or canonical variables are computed as an external module over all of the training samples, which is closer to the normal CCA computation process. 

The method of ACCAR is similar to the baseline method shown in figure 3. For the early M epochs of the whole training process, CCA transformation is computed every T epochs, which M and T are two hyperparameters.  The CCA transformations replace the weights of the fully connected layer of each stream, and the neural network train all layers include the fully connected layers with SGD or other batch-based optimization methods in the following T epochs, which means that the CCA transformations are initial weights for the FC-RGB and FC-Depth in figure 3 for every T epochs and the optimization along the direction which maximize the correlation. The objective function no longer needs to contain correlation as a regularization; the objective function of the baseline method is utilized for this structure. 

\subsubsection{Method 3: CCA Layer}

Except for two methods proposed by \cite{CCAR}, \cite{E2ECCA} also designed a neural network with CCA  for multi-modality information retrieval. The CCA computation of this method is almost the same with the first method, but the differences are no CCA related object functions are taken into the whole objective function, and the canonical variables or CCA projected feature vectors are inputs for the classifier. In our work, we modified this structure for image classification tasks based on the above baseline method, and the structure after modification is presented in the following figure 4. 

\begin{figure}[ht]
  \centering
    \includegraphics[width=0.7\paperwidth]{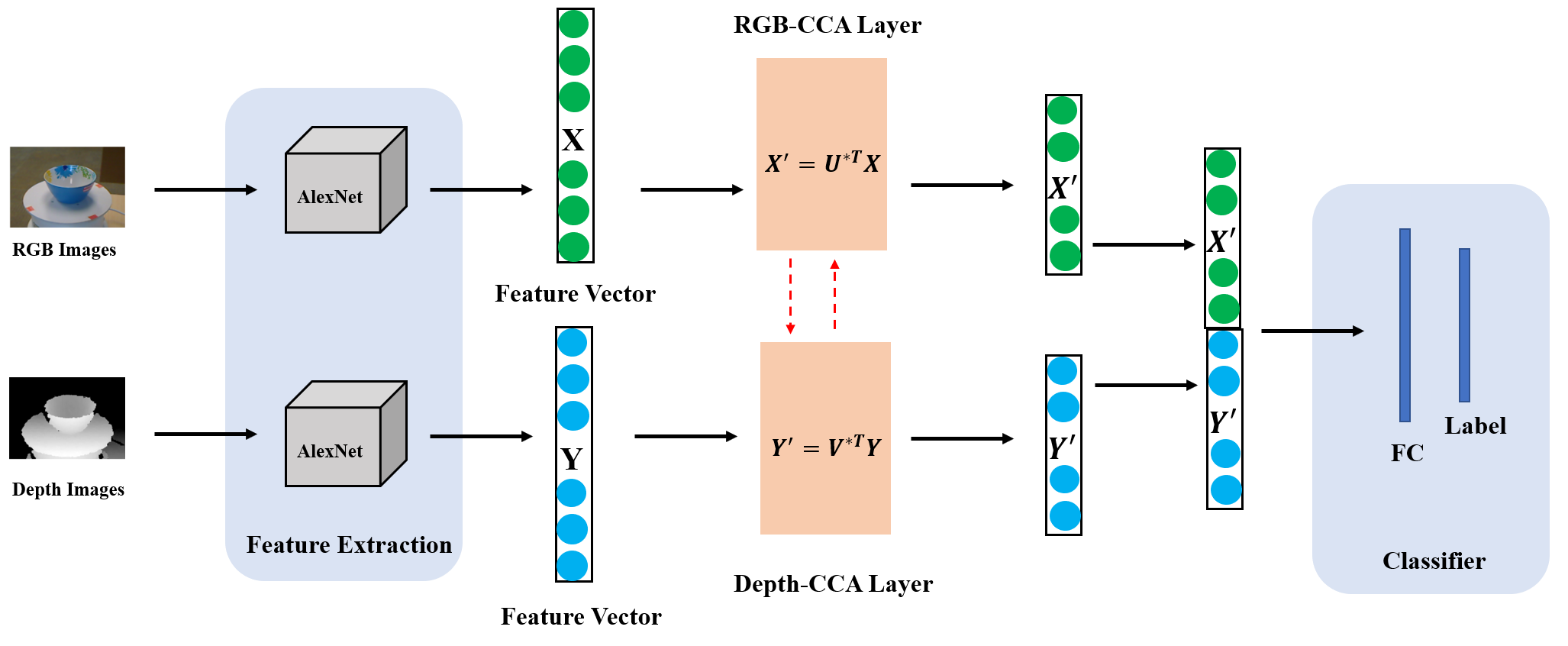}
  \caption{CCA Layer for Image Classification }
  \label{fig:network}
\end{figure}

The CCA layer calculates the transformations for every small-batch during training.  The output of each CCA layer is the projected feature of the corresponding view. Equation (3) is used for calculating canonical variables. As We mentioned above, the objective function is a cross-entropy loss for image classification, and it is a pairwise ranking loss in \cite{E2ECCA}.  For this method, the singularity problem over a small sample size still exists, so the relative big batch size is required.  

\subsection{Views Fusion For 2-D Object Detection}

The structures in the above section take feature vectors as inputs for the classifier,  so the output feature maps of feature extraction network are flattened, which means the spatial information of a feature map is lost.  However,  in many cases, spatial information contributes to higher classification accuracy or estimation accuracy, and even it is indispensable. Current 2-D object detection frameworks are among them. Therefore, we extend the ACCAR structure with two-dimensional CCA to the 2-D object detection task. 

\subsubsection{Two-Dimensional CCA (2DCCA)}

Two-dimensional canonical Correlation analysis was proposed by \cite{2DCCA}, which is utilized to analyze sets of variables, where each sample contains two-dimensional data. For two sets of image data, $X_t \in \mathbb{R} ^ {m_x \times n_x}$ and $Y_t \in \mathbb{R} ^ {m_y \times n_y}$, $t = 1,...,N$. 2DCCA aims to extraction one pair of transformation for each set of image data-one is named left transformation and the remaining one is right transformation so that the correlation between projected feature maps of two sets are maximum, which means:
\begin{equation}
argmax_{l_x, r_x, l_y, r_y} cov(\hat{X}, \hat{Y}) = argmax_{l_x, r_x, l_y, r_y} cov(l_x^TXr_x, l_y^TYr_y)
\end{equation}

The dimension of the left transformation of X and Y are $m_x \times d_1$ and $m_y \times d_1$, respectively. The dimension of the right transformation of X and Y are $n_x \times d_2$ and $n_y \times d_2$, respectively. After projections, the dimension of projected X and Y is $d_1 \times d_2$. 

Compared to the one-dimension CCA, there is no closed-form solution available for two-dimensional CCA.  Left transformations and the right transformation are calculated in an iteration way. When calculating $L_x$ and $L_y$, the $R_x$ and $R_y$ need to be fixed and vice versa. The following equations are used to calculate left transformations, and the computation of the right transformations is similar to this one.

\begin{equation}
\Sigma_{xx}^r = \frac{1}{N} \sum_{t=1}^N \tilde{X_t}R_xR_x^T\tilde{X}_t^T 
\end{equation}
\begin{equation}
\Sigma_{xy}^r = \frac{1}{N} \sum_{t=1}^N \tilde{X_t}R_xR_y^T\tilde{Y}_t^T
\end{equation}
\begin{equation}
\Sigma_{yx}^r = [\Sigma_{xy}^r]^T
\end{equation}
\begin{equation}
\Sigma_{yy}^r = \frac{1}{N} \sum_{t=1}^N \tilde{Y_t}R_yR_y^T\tilde{Y}_t^T
\end{equation}

\subsubsection{Baseline Method for 2-D Object Detection}

Mask RCNN \cite{Mask RCNN} is the baseline method we use for 2-D object detection, which is an advanced version of Faster R-CNN \cite{Faster RCNN} by adding one more regressor branch for mask prediction.  Three main parts of the Mask RCNN are-one feature extraction backbone that is consist of ResNet \cite{resnet} and Feature Pyramid Network (FPN) \cite{FPN}, and one Region Proposal Network, and one classier and regressor for estimation boxes and classification. ResNet \cite{resnet} generates feature maps with different dimensions, and FPN \cite{FPN} enhances feature maps from each level by adding the current feature map with the upsample feature map from the lower dimension. 

Multi-scales anchor boxes generated with each pixel as a center and each anchor box is labeled with negative, positive, or neutral. Labeling is based on the maximum overlapping area with any ground truth boxes, and the threshold for positive in our experiments is 0.7. The number of each type of anchor box is fixed to 33\% of the number of anchor boxes.  This label is a supervise signal to train RPN, and the predicted positive proposals are inputs for classifier and boxes estimation. 

As we introduce depth information as the extra stream to provide feature maps, we add one more feature extraction module for depth view and remove the mask estimation branch of Mask R-CNN \cite{Mask RCNN} because of lacking mask ground truth in our experiments. After modification, our baseline method for multi-views 2-D object detection can be expressed with the following figure 5. 

\begin{figure}[ht]
  \centering
    \includegraphics[width=0.7\paperwidth]{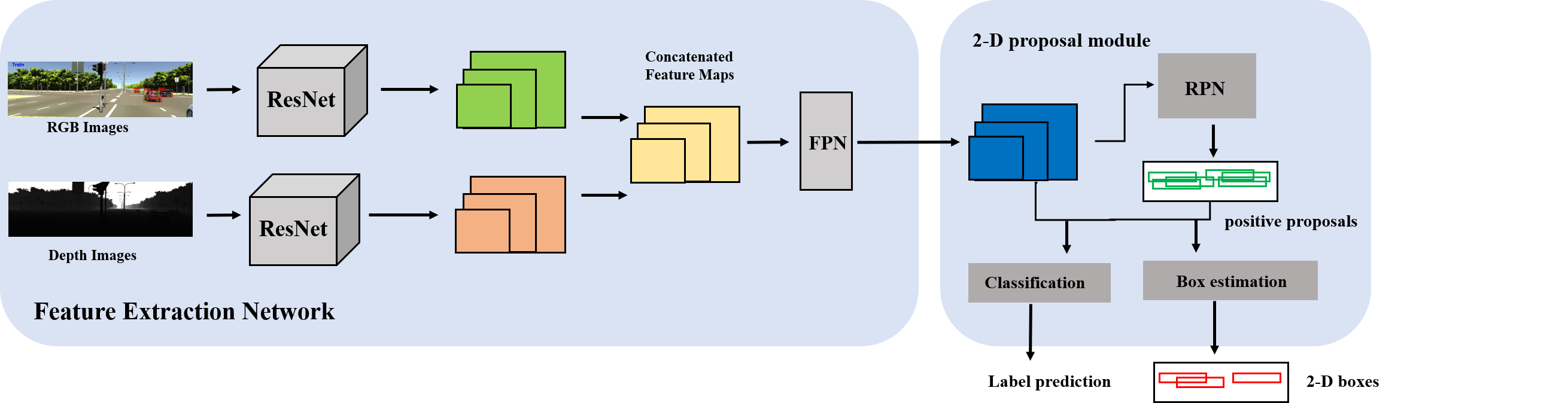}
  \caption{Baseline Method for Object Detection }
  \label{fig:network}
\end{figure}

\subsubsection{Extend ACCAR structure for 2-D Object Detection}

Based on the three structures for the image classification and the fact that the two-dimensional CCA is calculated in an iteration way,  the ACCAR structure is a possible method that can be combined with two-dimensional CCA for 2-D object detection \cite{2DCCA, CCAR}. Therefore, we extend the ACCAR structure to the 2DACCAR structure with two-dimensional CCA by adding a designed 2D-CCA  layer before FPN. Figure 6 shows the 2DACCAR structure.     

\begin{figure}[ht]
  \centering
    \includegraphics[width=0.7\paperwidth]{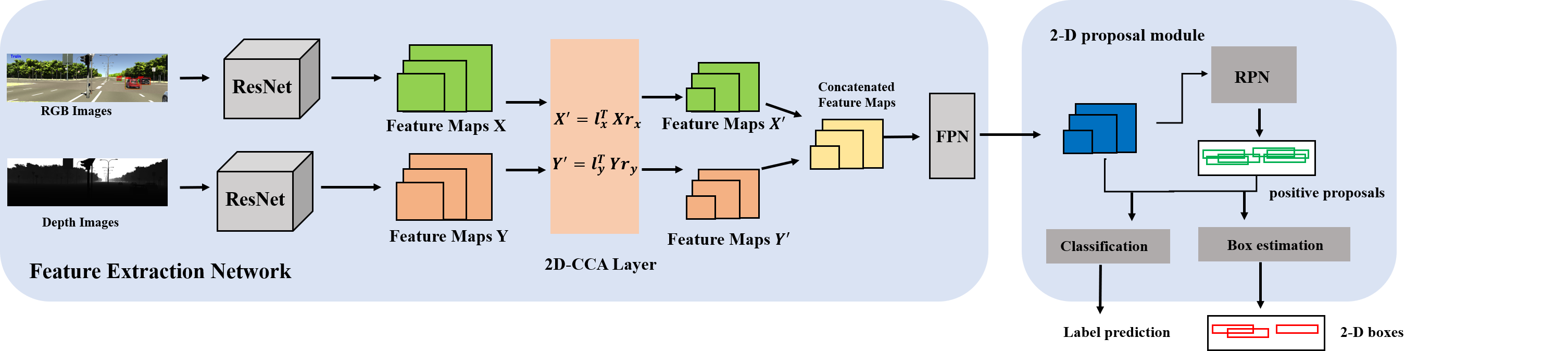}
  \caption{2DACCAR for Object Detection}
  \label{fig:network}
\end{figure}

As the figure shows, the 2D-CCA layer takes the feature maps from the RGB view and the depth view as inputs. The outputs are feature maps with a smaller size for each view independently, and the size of an output feature map is a hyperparameter, which is equal to the half size of an original image in our experiments. Four parameters of a 2D-CCA layer stand for two left transformations and two right transformations in section 3.2.1, and these parameters are initialized by uniform distribution during training. For every T epochs, which T is the 2DCCA calculation frequency, the external 2DCCA module calculates two pairs of 2DCCA transformation, and the parameters are replaced with two pairs of transformation.  Because an input image is corresponding to several feature maps, we compute the mean over all of the feature maps and use the mean feature map as inputs for the external 2D-CCA module. Due to multi-dimension feature maps required for detecting objects with distinguishing sizes,  one 2D-CCA layer is added for one pair of feature maps from the  RGB and depth view with the same dimension, which means five 2D-CCA layers are added if five different dimensions of feature maps required. 

After getting the 2D-CCA projected feature maps for each view, the feature maps of two views are concatenated as inputs for the FPN, and the following modules are remaining to be same. 

\subsection{Depth Estimation}

We use the DenseDepth \cite{DDepth} as our external depth estimation module, and it only requires a single monocular image as input for reasonable depth estimation without using monocular images in sequence or stereo images. That is why it is a very considerable method for depth estimation in our works because many 2-D object detection benchmarks provide no videos or stereo images.   

The DenseDepth estimation network is implemented in an encoder-decoder manner, and the encoder part is based on the DenseNet. The idea of the DenseNet is using the features maps from all the preceding convolutional layers so that the final feature maps will contain all information from every convolutional layer. The decoder part of dense depth is similar to the encoder part. It consists of several convolutional blocks, and the input of every block is a combination of the output of the preceding block and corresponding size feature maps in the encoder.

The loss function of this network contains three sub-loss functions. One loss is point-wise L1 loss defined on the predicted depth and ground truth depth, which is shown in equation 12. 

\begin{equation}
    L_{depth}(y, \hat{y}) = \frac{1}{n}\sum_p^n |y_p - \hat{y_p}|
\end{equation}

The $\hat{y}$ means the predicted depth images, and it calculates the difference over every pixel. The second loss is the gradient loss, which means whether the pixels in the predicted depth images have the same direction to increase or decrease with the ground truth depth images. 

\begin{equation}
    L_{grad}(y, \hat{y}) = \frac{1}{n} \sum_p^n |g_x(y_p, \hat{y_p})| + |g_y(y_p, \hat{y_p})|
\end{equation}
    
\begin{equation}
     L(y, \hat{y}) = \lambda  L_{depth}(y, \hat{y}) +  L_{grad}(y, \hat{y}) + \frac{1-SSIM(y, \hat{y})}{2}
\end{equation}

Equation 14 shows the objective function of DenseDepth, and the $SSIM(y, \hat{y})$ means Structural Similarity, which is commonly used for image reconstruction task. 
\section{Experiments}
\subsection{Datasets and Input Preprocessing}
\subsubsection{Washington RGBD}

It has 51 categories in total, and for each category, which contains at least three objects and each object is captured from different angles \cite{WRGBD}. The frames for one object are continuous. Because the object task is image recognition, only one frame for every continuous five frames is picked out as a training or validation image. Two different objects are selected from each category and assigned to the validation set and the test set, respectively. No frames are covered by more than one subset. 

For compatible with the original input image size of AlexNet, the images are scaled and padded into $227 \times 227$ without deforming them. First, the longer side of an input image is upscaled to 227 or downscaled to 227, and the shorter side is padded to 227 with zeros. For the preprocessing of a ground truth depth image, the pixels with missing data are filled by the mean of adjacent pixels in $5 \times 5$ grid. Furthermore, surface normal images with three channels are calculated based on the depth images after filling missing data to use pre-trained models of AlexNet on Imagenet fully. 

\subsubsection{Virtual KITTI}

This dataset contains synthetic frames in sequence from KITTI \cite{kitti}. There are five different scenes, and each scene contains ten different conditions in terms of viewpoints and light conditions. The original $1242 \times 375$ pixel images are scaled and padding into $768 \times 768$ because the size of input images should be multiple of 64 to downsample, and our machines for experiments have limited memory for bigger sizes. Five thousand images are chosen randomly over all of the scenes and conditions; the rest of the images are used for validation and inferencing. 

For the depth prediction task, the frames are scaled and padded into $1280 \times 384$ for downsampling. As the output depth information of DenseDepth is the half size of the input monocular images, the predicted depth images are rescaled and unpadded into  $1242 \times 375$ after depth estimation.

\subsubsection{KITTI and CityScape}

The frames of KITTI \cite{kitti} and CItyScape \cite{cscape} are used for quality analysis by estimating the depth information from RGB images and perform the 2-D object detection on RGB and depth images. The frames of raw KITTI dataset have the same resolution with frames of Virtual KITTI, and the scaling and padding are also required for inferencing for these two datasets. The upsampling mentioned in the above section for predicted depth information is also unavoidable for these two datasets.  As no depth images available, and the converted depth images from LiDAR points cloud are very sparse,  no re-training for KITTI and CityScape. 

\subsection{Implement Details}
\subsubsection{Experimental Setup}
All the image classification experiments are executed on a desktop with Ubuntu 18.04 LTS system, Intel Xeon 16 cores CPU, Quadro P400 with 8 GB memory, and 64 GB memory. For the experiments of 2-D object detection,  Quadro K6000 with 12 GB memory is used and keep other configurations are same.

\subsubsection{Implementation of Image Classification}

Before training three methods mentioned in section 3.1,  we first trained three single-view models for RGB,  Depth, and Surface Normal separately. The structure for single-view training can be constructed by removing the useless feature extraction branch.      The structure for single-view training can be constructed by removing the useless feature extraction branch. The training for RGB images loaded the pre-trained model of AlexNet on ImageNet, and the training of depth and surface normal images use uniform distribution to initialize the parameters. 

We trained two-views models on two data combination- RGB images with Depth images, and RGB images with Surface Normal Images.  The parameters of layers in the AlexNet are initialized by the single-view pre-trained model for each view, , and other layers are initialized by uniform distribution initializer.  For the ACCAR method, the model is trained with 80 epochs, and CCA happens in the first 60 epochs. The CCA calculation frequency is set to 10 and 20 separately. In terms of CCA regularization structure, multiple values of lambda are involved in analyzing the effect of the CCA objective function. As the CCA Layer method cannot converge in our structure, we are not listing any results and comparison of this method. We think the reason is the canonical variables on few samples cannot express the correlation over whole training samples accurately. 
\subsubsection{Implementation of 2-D Object Detection}

The dataset involved for the 2-D object detection task is Virtual KITTI, and the split of training set we have mentioned in section 4.1.2. As we did in the image classification experiments, we first trained with RGB images and ground truth depth images independently by load the pre-trained model of Mask RCNN on the COCO dataset \cite{coco}. Since the depth images are one channel, we did not load the weights of the first convolution layer. These two models are trained with 15 epochs. Two-views structure loads these two pre-trained models for the corresponding view in terms of feature extraction branch. However, the initial weights of the RPN and the classifier are from the pre-trained model of RGB images. The two-views structure without  CCA is trained with 30 epochs. 

The weights initializing of the two-dimensional ACCAR method are the same as the structure without two-dimensional CCA. There are two separate training strategies for the two-dimensional ACCAR structure. First is fine-tuning 15 epochs with the custom 2D-CCA layer and then calculating CCA every five epochs until 30 epochs are done. Another one is calculating CCA every ten epochs until 30 epochs are done. We named these two ACCAR strategies 2DACCAR-1 and 2DACCAR-2, respectively, in the following results section. 

\subsection{Results of Image Classification}

Table 1 shows the overall performance of all methods with two views involved in our experiments, and the $\lambda$ is set to 0.5 for CCAR, and the frequency of CCA calculation for ACCAR is set to 20 in this table. For all methods, the combination of RGB images and surface normal images achieves higher classification accuracy,  and because of no overlapping frames for validation set and test set, the validation set seems more challenge than the test set, and three methods obtain higher accuracy on the test set.  For the two structures with CCA, they both cannot achieve better performance than the baseline method, and the classification accuracy of the ACCAR method is close to the results of the baseline method.  The results of [3] are much better even for the baseline method, but the author did not release their split method for the training set and test set,  we cannot make a complete comparison. 

\begin{table}[ht]
 \caption{Performance Over Methods}
  \centering
  \begin{tabular}{lllll}
    \toprule
    \multirow{2}{*}{Methods} & \multicolumn{2}{c}{Validation Set} &  \multicolumn{2}{c}{Test Set}\\
    \cline{2-5} 
    &RGB-D     & RGB-SN & RGB-D     & RGB-SN\\
    \midrule
    Baseline & 70.68  & 78.22 & 78.94 & 81.77     \\
    CCAR     & 68.73 & 69.36 & 72.31 & 74.86      \\
    ACCAR     & 70.42 & 77.02 & 76.50 & 81.83  \\
    \bottomrule
  \end{tabular}
  \label{tab:table}
\end{table}

Table 2 shows the results of three different models on two different subsets, and use the same pre-trained model as table 1, and the $\lambda$ is set to 0.3 for CCAR, and the frequency of CCA calculation for ACCAR is set to 20. We find that the accuracy of the CCAR method dropped a lot for subsets that are not used for training, which suggests that drawback of calculating CCA over a small batch of data. 
\begin{table}[ht]
 \caption{Performance Over Methods with CCA on multi subsets}
  \centering
  \begin{tabular}{lllllllll}
    \toprule
    \multirow{2}{*}{Methods} & \multicolumn{2}{c}{Validation Set2} &  \multicolumn{2}{c}{Test Set2} &\multicolumn{2}{c}{Validation Set2} & \multicolumn{2}{c}{Test Set2}\\
    \cline{2-9} 
    &RGB-D     & RGB-SN & RGB-D     & RGB-SN&RGB-D     & RGB-SN&RGB-D     & RGB-SN\\
    \midrule
    CCAR     & 44.21 &  45.82 & 37.81 & 38.03 & 44.40& 45.20 & 36.95 & 38.18      \\
    ACCAR     & 70.54 & 77.05 & 74.82 & 81.40 & 70.16 & 75.78 & 76.12 & 82.60  \\
    \bottomrule
  \end{tabular}
  \label{tab:table}
\end{table}

Several results of the performance of CCAR with multiple a are shown in table 3, and no exact pattern can be inferred from these results, and the increasing and decreasing of the accuracy are random. The situation is the same in terms of the CCA calculation frequency of the ACCAR method.  The results of calculating CCA every ten epochs and twenty epochs are very close. 
\begin{table}[h!]
 \caption{Performance of the CCAR with multiple $\lambda$}
  \centering
  \begin{tabular}{lllll}
    \toprule
    \multirow{2}{*}{$\lambda$} & \multicolumn{2}{c}{Validation Set} &  \multicolumn{2}{c}{Test Set}\\
    \cline{2-5} 
    &RGB-D     & RGB-SN & RGB-D     & RGB-SN\\
    \midrule
    0.1 & 70.65  & 67.70 & 76.36 & 75.07     \\
    0.3     & 69.06 & 69.00 & 71.57 & 73.89      \\
    0.5     & 68.73 & 69.36 & 72.31 & 74.86  \\
    0.7     & 69.70 & 69.73 & 72.37 & 73.45  \\
    \bottomrule
  \end{tabular}
  \label{tab:table}
\end{table}

\begin{table}[h!]
 \caption{Performance pf the ACCAR with different CCA calculation frequency}
  \centering
  \begin{tabular}{lllll}
    \toprule
    \multirow{2}{*}{CCA Frequency} & \multicolumn{2}{c}{Validation Set} &  \multicolumn{2}{c}{Test Set}\\
    \cline{2-5} 
    &RGB-D     & RGB-SN & RGB-D     & RGB-SN\\
    \midrule
    10 & 70.50 & 78.18 & 76.36 & 81.85     \\
    20 & 70.42 & 77.02 & 76.50 & 81.83    \\
    \bottomrule
  \end{tabular}
  \label{tab:table}
\end{table}

\subsection{Results of 2-D Object Detection}

Table 5 shows the results of each method with the RGB images and ground truth depth images as inputs. All the pre-trained models of these methods are trained by RGB images and ground truth depth images. "Layer-Only" means the structure with the 2D-CCA layers but no 2DCCA calculation and no parameter replacement. 
Three evaluation metrics for quantitative analysis are mAP, mRecall, and mIoU. The mAP is the mean of precision over all classes, and mRecall is the average of the percent that ground truth boxes are predicted, and mIoU is the mean of the overlapping area of all inferencing images. When taking the ground truth depth images as inputs, the baseline method achieves better performance for all three evaluation metrics because, except the baseline method, all other methods use smaller feature maps as input for classifier and regressor, which means some information lost. However, every pixel in ground truth depth, especially for Virtual KITTI, contains the accurate depth value, so the lost information is useful.

\begin{table}[h!]
 \caption{Results with ground truth depth and RGB images as input}
  \centering
  \begin{tabular}{llllllll}
    \toprule
    \multirow{2}{*}{Methods} & \multicolumn{3}{c}{Validation Set} &  \multicolumn{3}{c}{Test Set}\\
    \cline{2-7} 
    &mAP     & mRecall & mIoU     &mAP     & mRecall & mIoU\\
    \midrule
    Baseline & 98.01 & 96.69 & 91.31 & 98.29 & 97.18 & 91.80     \\
    Layer-Only & 95.57 & 92.84 & 86.40 & 97.75 & 95.15 & 88.68    \\
    2DACCAR-1 & 97.29 & 94.11 & 88.31 & 98.02&95.60&88.87     \\
    2DACCAR-2 & 95.87 & 92.95 & 84.65 & 96.39 & 90.28 & 84.74   \\
    \bottomrule
  \end{tabular}
  \label{tab:table}
\end{table}

Table 6 shows the results when taking RGB images and predicted depth images as inputs with the pre-trained model on RGB images and ground truth depth images. The performance of the baseline method drops a lot for this situation, which suggests that the 2D-CCA layer reduces the noisy information inside the feature maps. The following images shows the visual differences between methods. 

\begin{table}[h!]
 \caption{Results with predicted depth images and RGB images as input}
  \centering
  \begin{tabular}{llllllll}
    \toprule
    \multirow{2}{*}{Methods} & \multicolumn{3}{c}{Validation Set} &  \multicolumn{3}{c}{Test Set}\\
    \cline{2-7} 
    &mAP     & mRecall & mIoU     &mAP     & mRecall & mIoU\\
    \midrule
    Baseline & 70.41 & 66.94 & 75.98 & 69.89 & 66.50 & 75.26     \\
    Layer-Only & 97.67 & 92.71 & 85.80 & 97.39 & 92.87 & 85.93    \\
    2DACCAR-1 & 98.14 & 96.36 & 87.57 & 97.91&96.10&87.57     \\
    2DACCAR-2 & 96.61 & 90.34 & 84.32 & 96.44 & 89.89 & 84.26   \\
    \bottomrule
  \end{tabular}
  \label{tab:table}
\end{table}

Figure 7-12 show the 2-D predicted boxes of Virtual KITTI, KITTI, and CityScape with pre-trained models from baseline method and 2DACCAR method. Compared to the 2DACCAR, baseline method misses obvious instances.

\subsection{Discussion and Further Works}
Although the current 2DACCAR structure obtains better performance when using the predicted depth images and RGB images as inputs, the network is trained on ground truth depth images. Based on our designed pipeline for 3-D object detection, predicted depth images are used for training the 2-D proposal network and 3-D proposal network. Therefore, we will use predicted depth images for training in further works. 

The smaller feature maps are a possible reason for poor performance when the inputs are ground truth images and RGB images, even if the information included in the feature maps already enough for reasonable results. Therefore, our further work will focus on a better concatenating method without reducing the size of feature maps. 

Furthermore, our proposed 3-D object detection pipeline heavily depends on the performance of the 2-D proposal network, which means the more accurate results of 2-D proposals will boost the whole framework. We will aim to use the 3-D segmentation over points cloud to improve the performance of 2-D proposals.

\begin{figure}[htb!]
  \begin{subfigure}[b]{1\linewidth}
    \centering
    \includegraphics[width=1\linewidth]{70_two_view.png}
  \end{subfigure}
  \begin{subfigure}[b]{1\linewidth}
  \centering
    \includegraphics[width=1\linewidth]{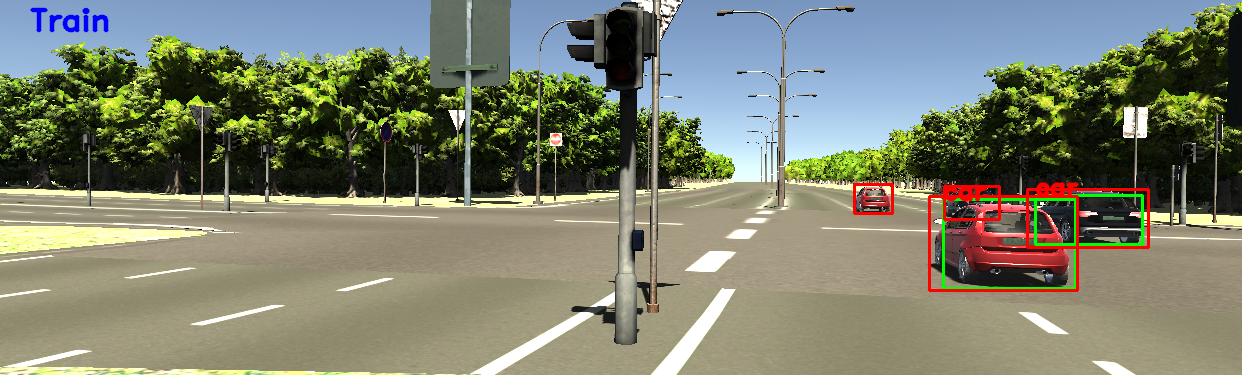}
  \end{subfigure}
  \begin{subfigure}[b]{1\linewidth}
  \centering
    \includegraphics[width=1\linewidth]{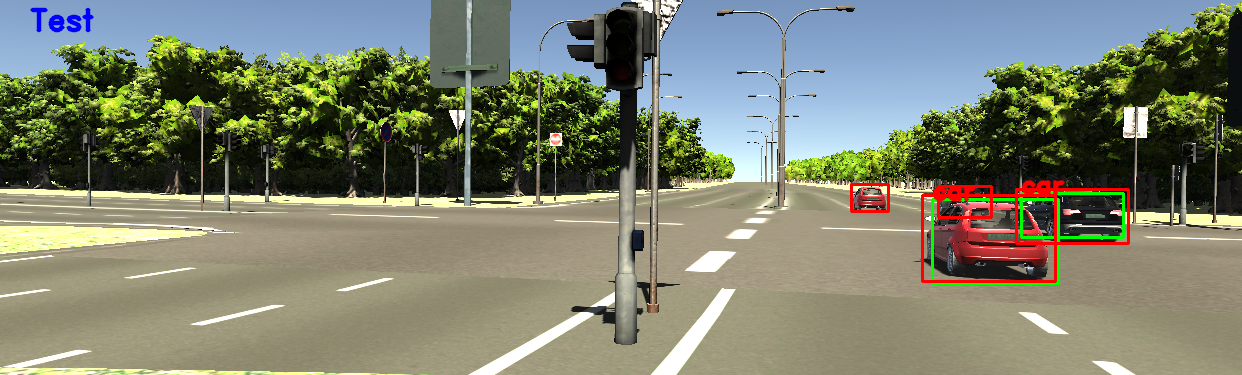}
  \end{subfigure}
  \begin{subfigure}[b]{1\linewidth}
  \centering
    \includegraphics[width=1\linewidth]{72_two_view.png}
  \end{subfigure}
  \caption{\small Four continuous output frames of baseline method taking predicted depth images and RGB images as inputs}
  \label{fig:image_group_1}
\end{figure}

\begin{figure}[h!]
  \begin{subfigure}[b]{1\linewidth}
    \centering
    \includegraphics[width=1\linewidth]{70_accar.png}
  \end{subfigure}
  \begin{subfigure}[b]{1\linewidth}
  \centering
    \includegraphics[width=1\linewidth]{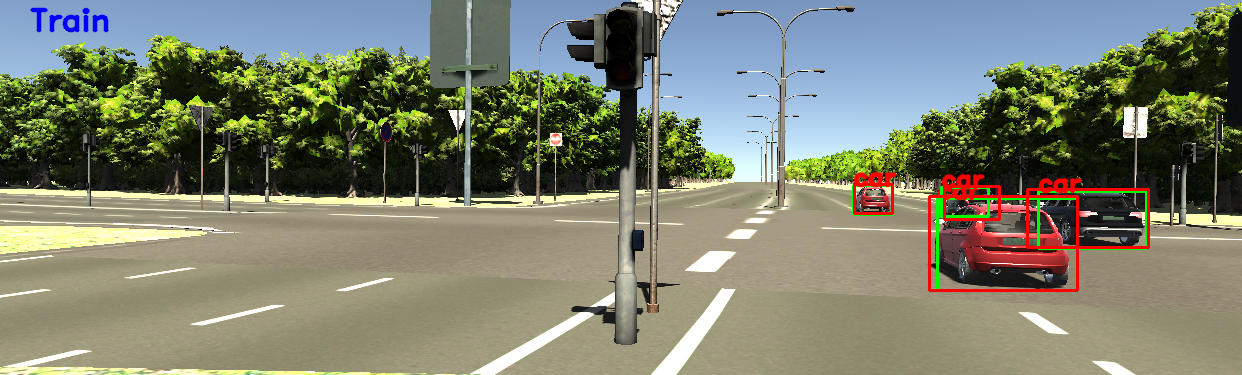}
  \end{subfigure}
  \begin{subfigure}[b]{1\linewidth}
  \centering
    \includegraphics[width=1\linewidth]{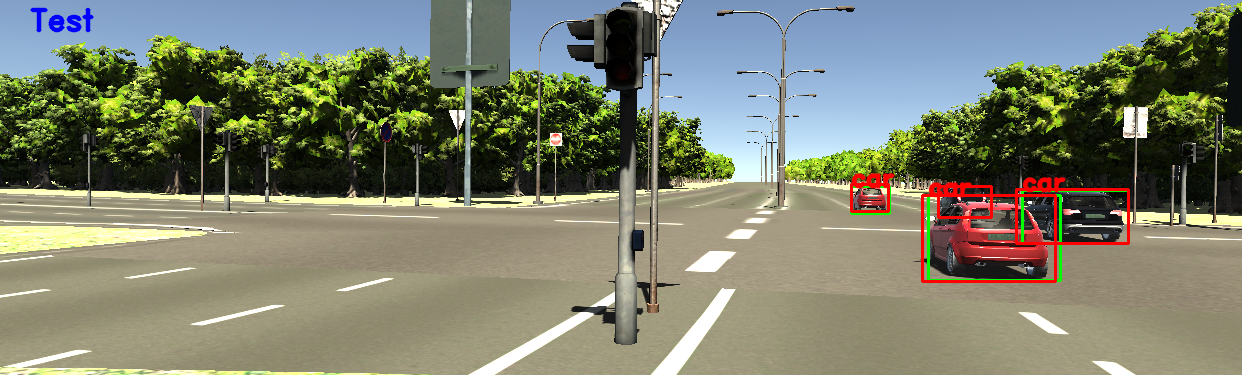}
  \end{subfigure}
  \begin{subfigure}[b]{1\linewidth}
  \centering
    \includegraphics[width=1\linewidth]{72_accar.png}
  \end{subfigure}
  \caption{\small Four continuous output frames of 2DACCAR-1 taking predicted depth images and RGB images as inputs}
  \label{fig:image_group_1}
\end{figure}

\begin{figure}[h!]
  \begin{subfigure}[b]{0.8\linewidth}
    \centering
    \includegraphics[width=0.8\linewidth]{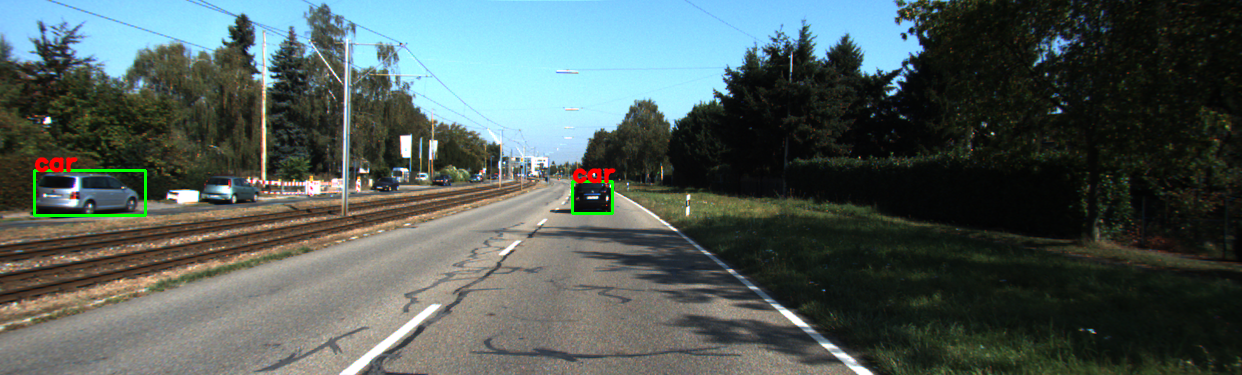}
  \end{subfigure}
  \begin{subfigure}[b]{0.8\linewidth}
  \centering
    \includegraphics[width=0.8\linewidth]{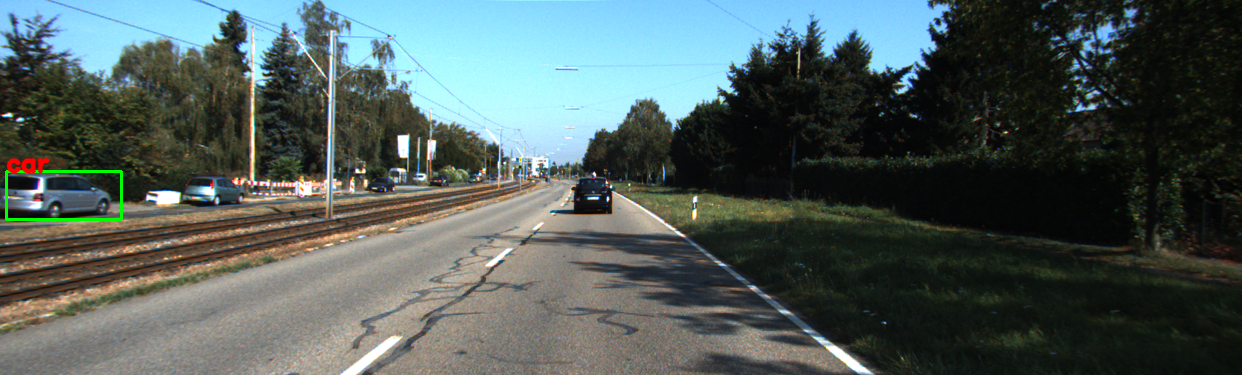}
  \end{subfigure}
  \begin{subfigure}[b]{0.8\linewidth}
  \centering
    \includegraphics[width=0.8\linewidth]{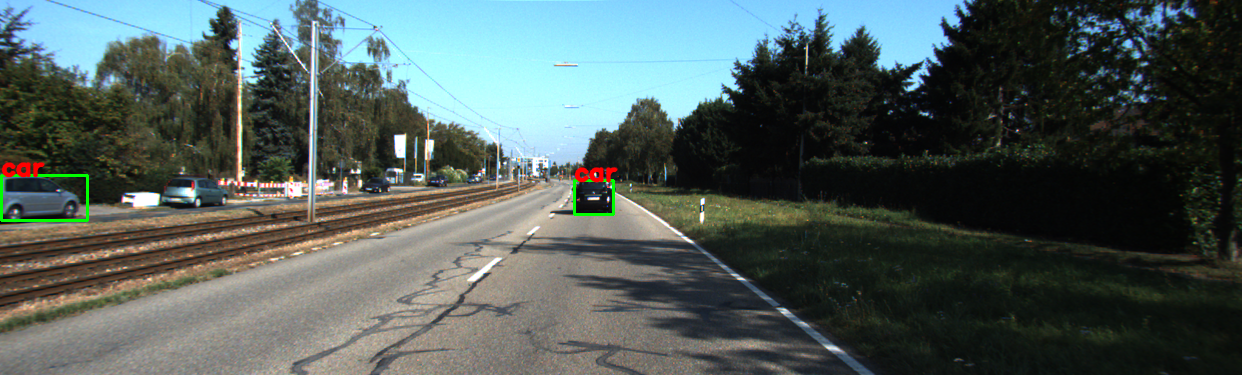}
  \end{subfigure}
  \caption{\small Four continuous output frames of baseline method taking predicted depth images and RGB images from KITTI as inputs}
  \label{fig:image_group_1}
\end{figure}

\begin{figure}[h!]
  \begin{subfigure}[b]{0.8\linewidth}
    \centering
    \includegraphics[width=0.8\linewidth]{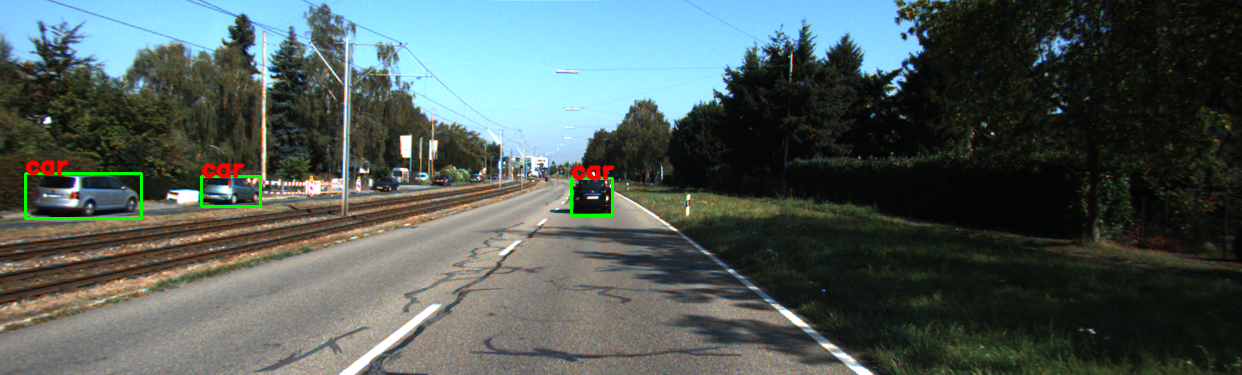}
  \end{subfigure}
  \begin{subfigure}[b]{0.8\linewidth}
  \centering
    \includegraphics[width=0.8\linewidth]{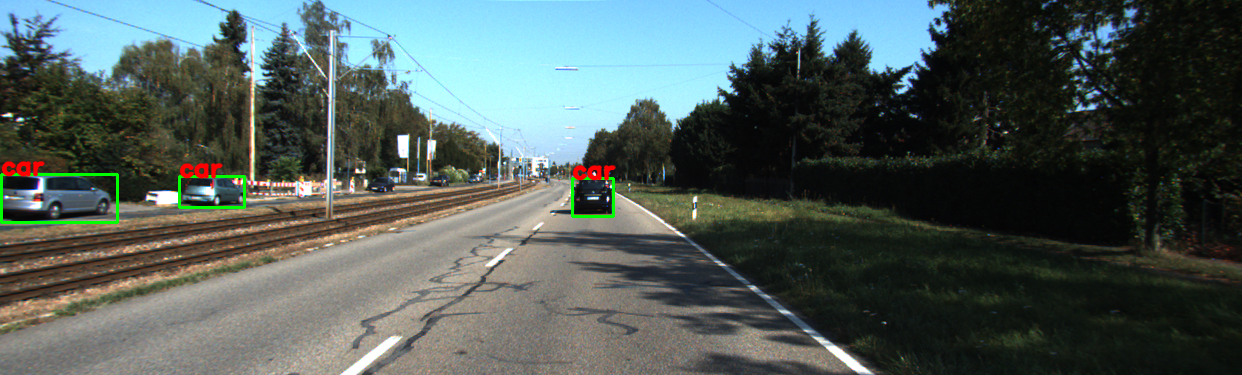}
  \end{subfigure}
  \begin{subfigure}[b]{0.8\linewidth}
  \centering
    \includegraphics[width=0.8\linewidth]{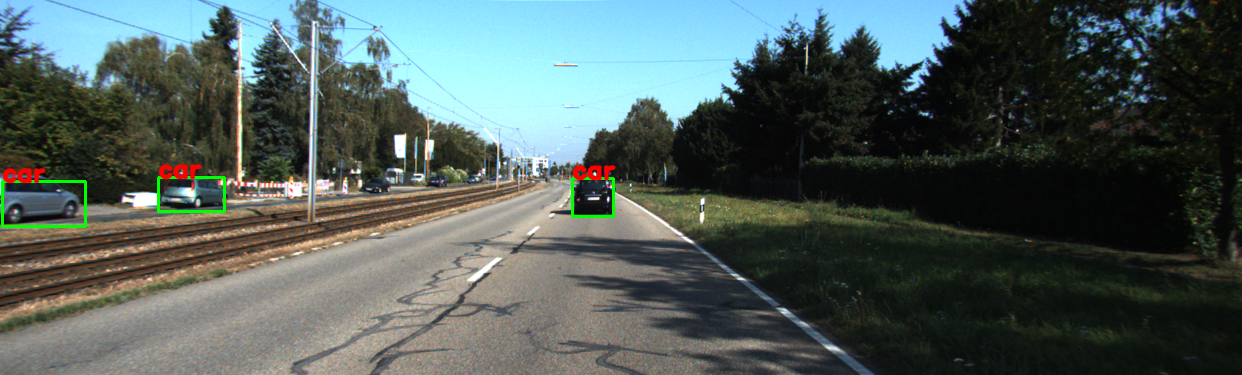}
  \end{subfigure}
  \caption{\small Three continuous output frames of 2DACCAR-1 taking predicted depth images and RGB images from KITTI as inputs}
  \label{fig:image_group_1}
\end{figure}

\begin{figure}[h!]
  \begin{subfigure}[b]{0.8\linewidth}
    \centering
    \includegraphics[width=0.8\linewidth]{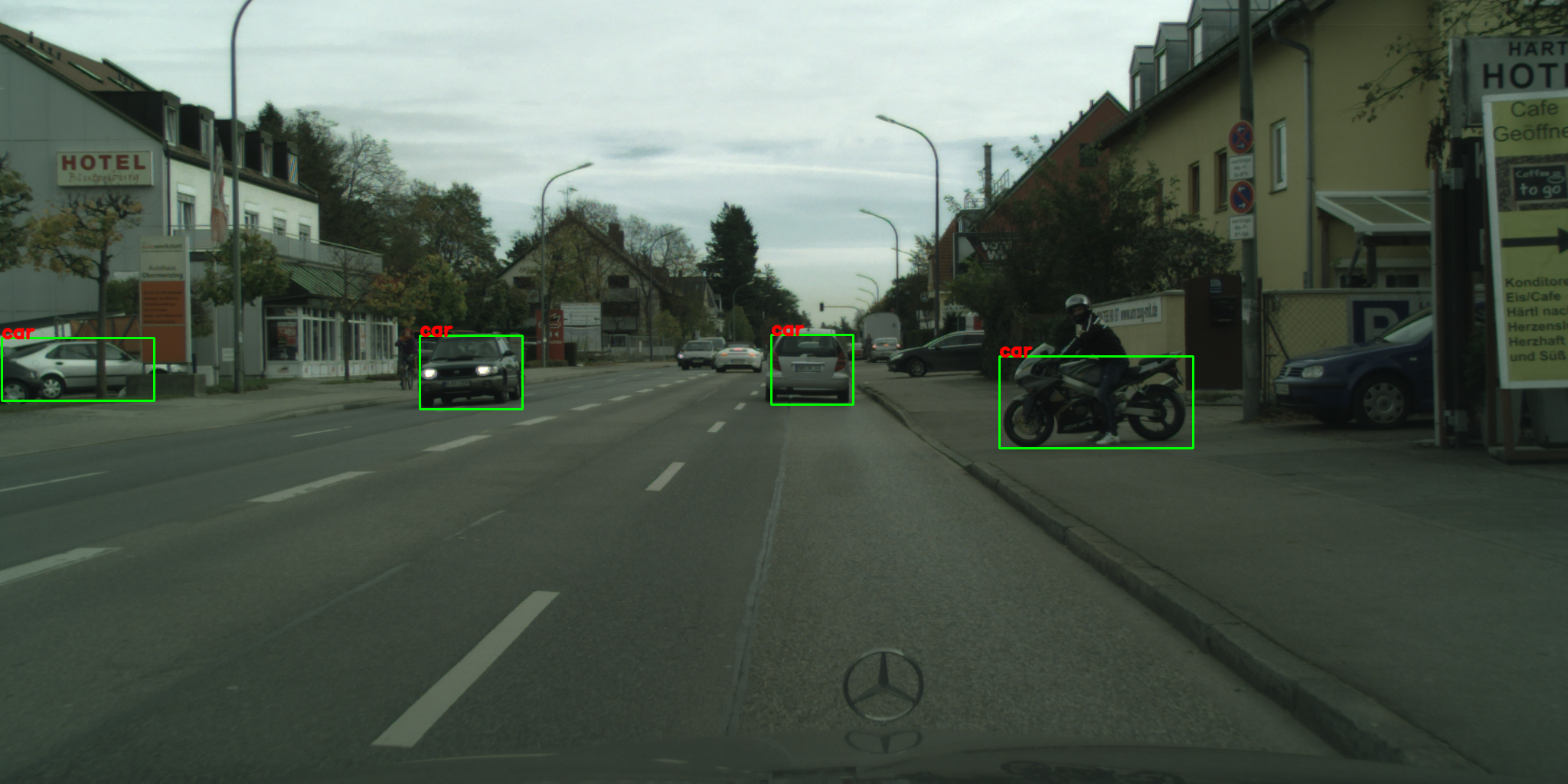}
  \end{subfigure}
  \begin{subfigure}[b]{0.8\linewidth}
  \centering
    \includegraphics[width=0.8\linewidth]{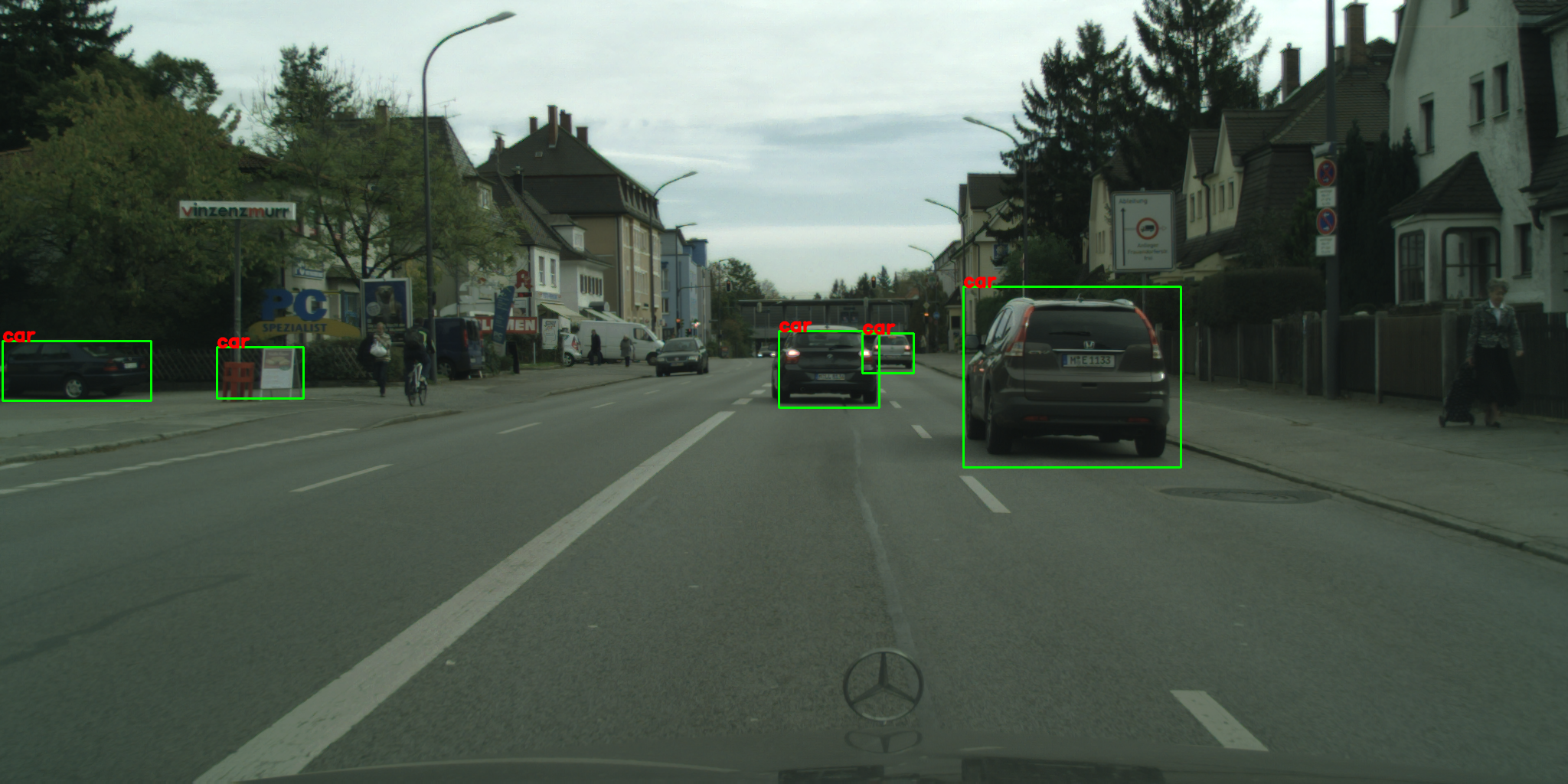}
  \end{subfigure}
  \begin{subfigure}[b]{0.8\linewidth}
  \centering
    \includegraphics[width=0.8\linewidth]{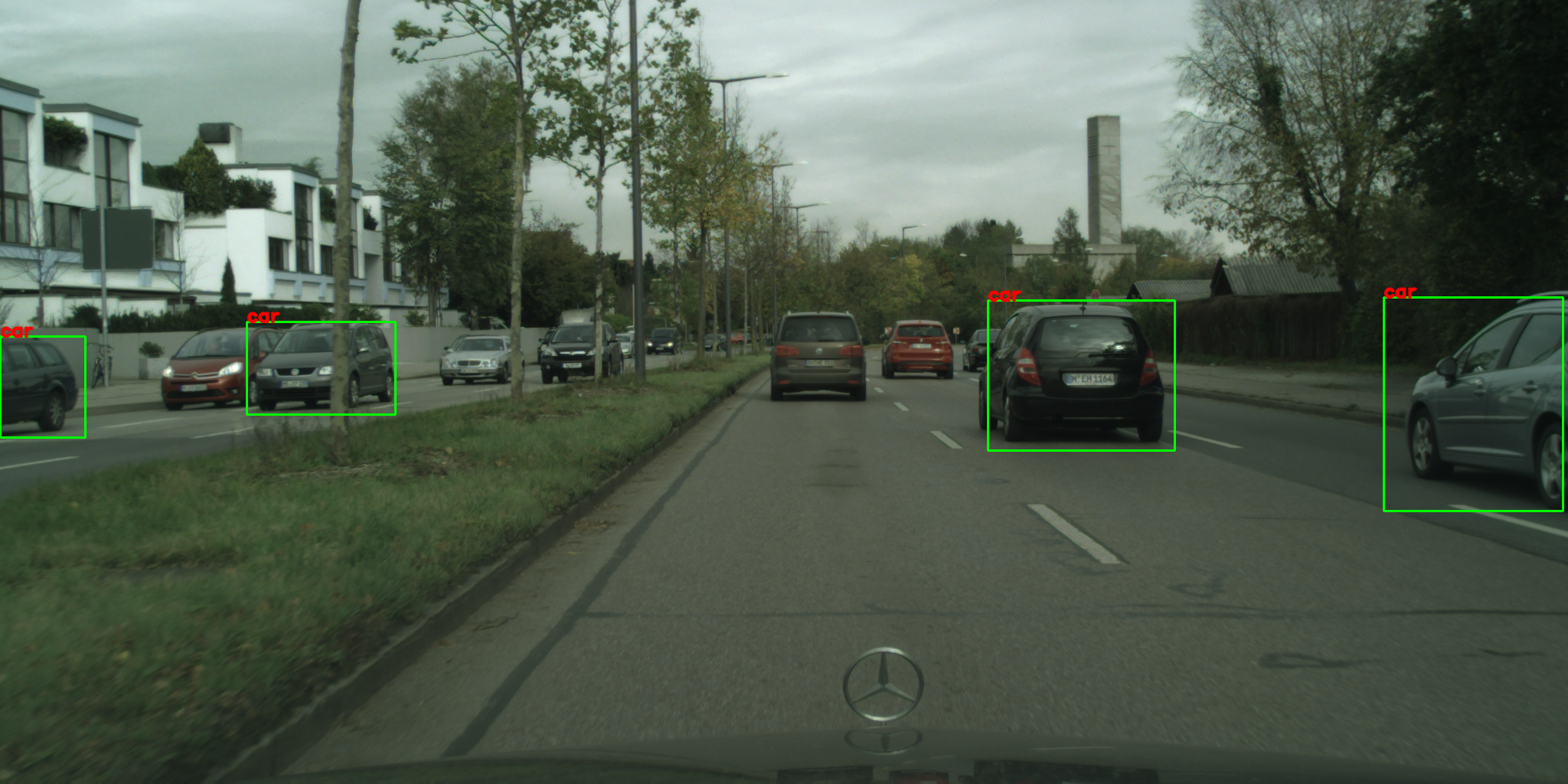}
  \end{subfigure}
  \caption{\small Four continuous output frames of baseline method taking predicted depth images and RGB images from CityScape as inputs}
  \label{fig:image_group_1}
\end{figure}

\begin{figure}[h!]
  \begin{subfigure}[b]{0.8\linewidth}
    \centering
    \includegraphics[width=0.8\linewidth]{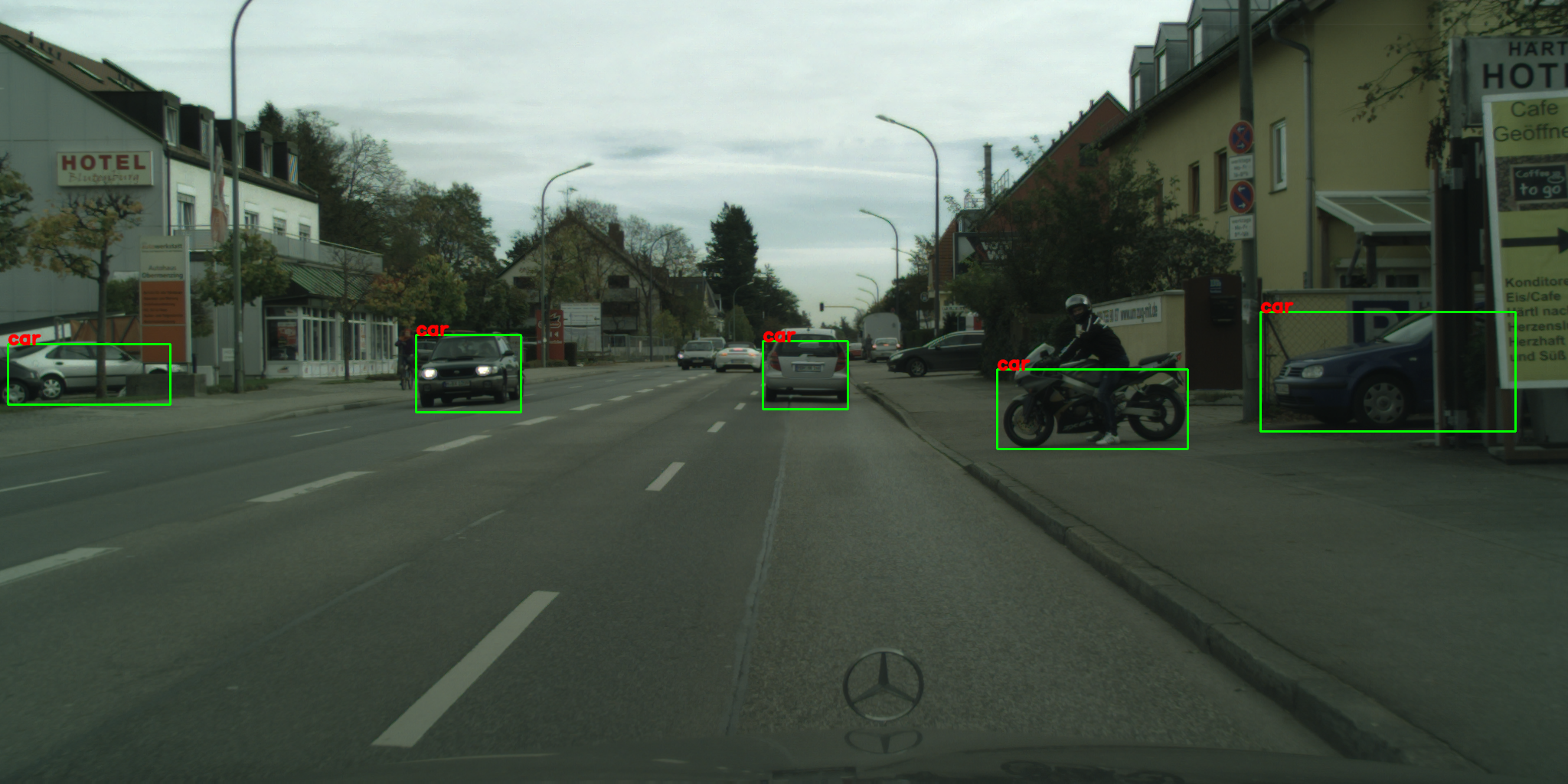}
  \end{subfigure}
  \begin{subfigure}[b]{0.8\linewidth}
  \centering
    \includegraphics[width=0.8\linewidth]{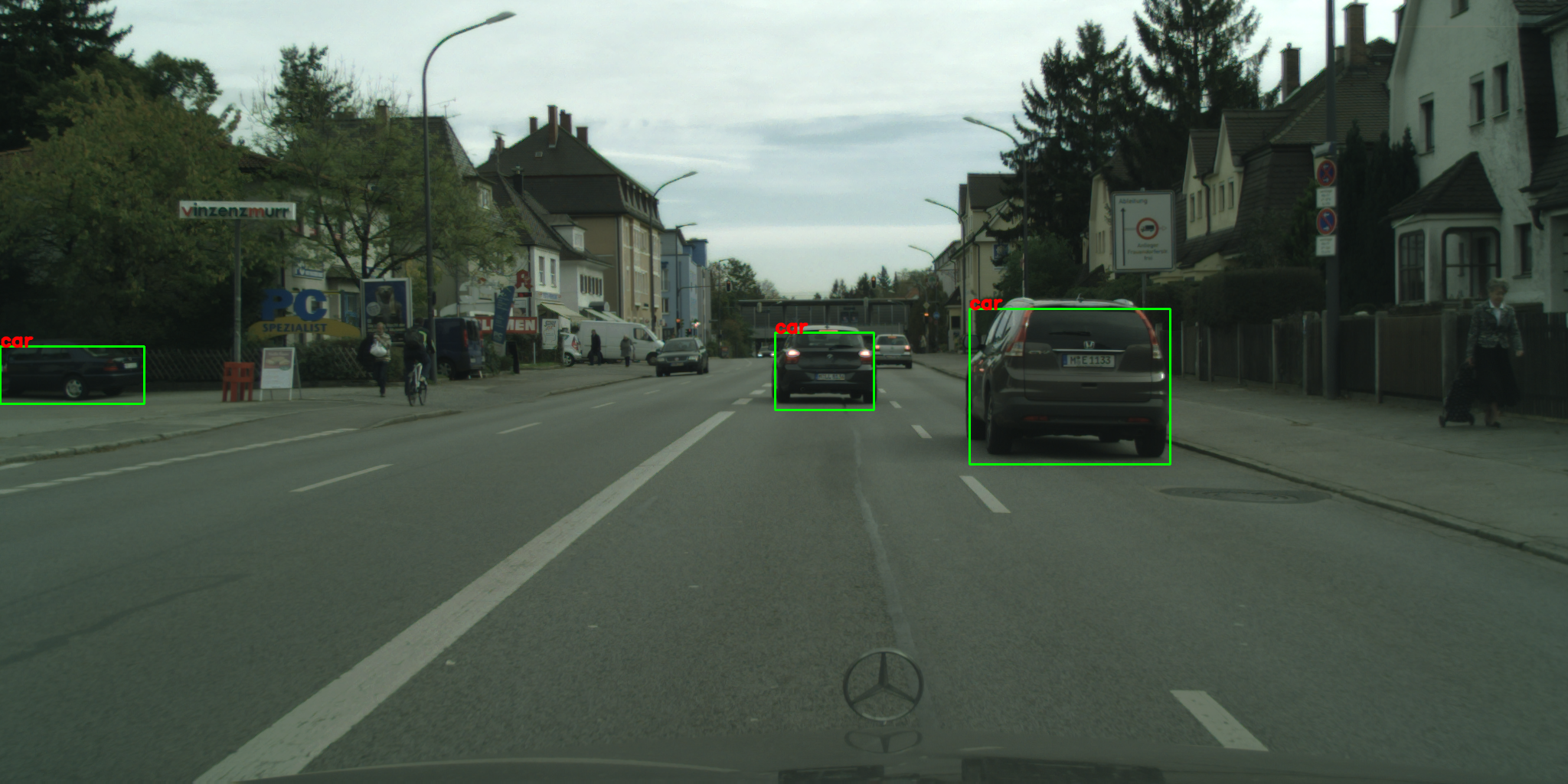}
  \end{subfigure}
  \begin{subfigure}[b]{0.8\linewidth}
  \centering
    \includegraphics[width=0.8\linewidth]{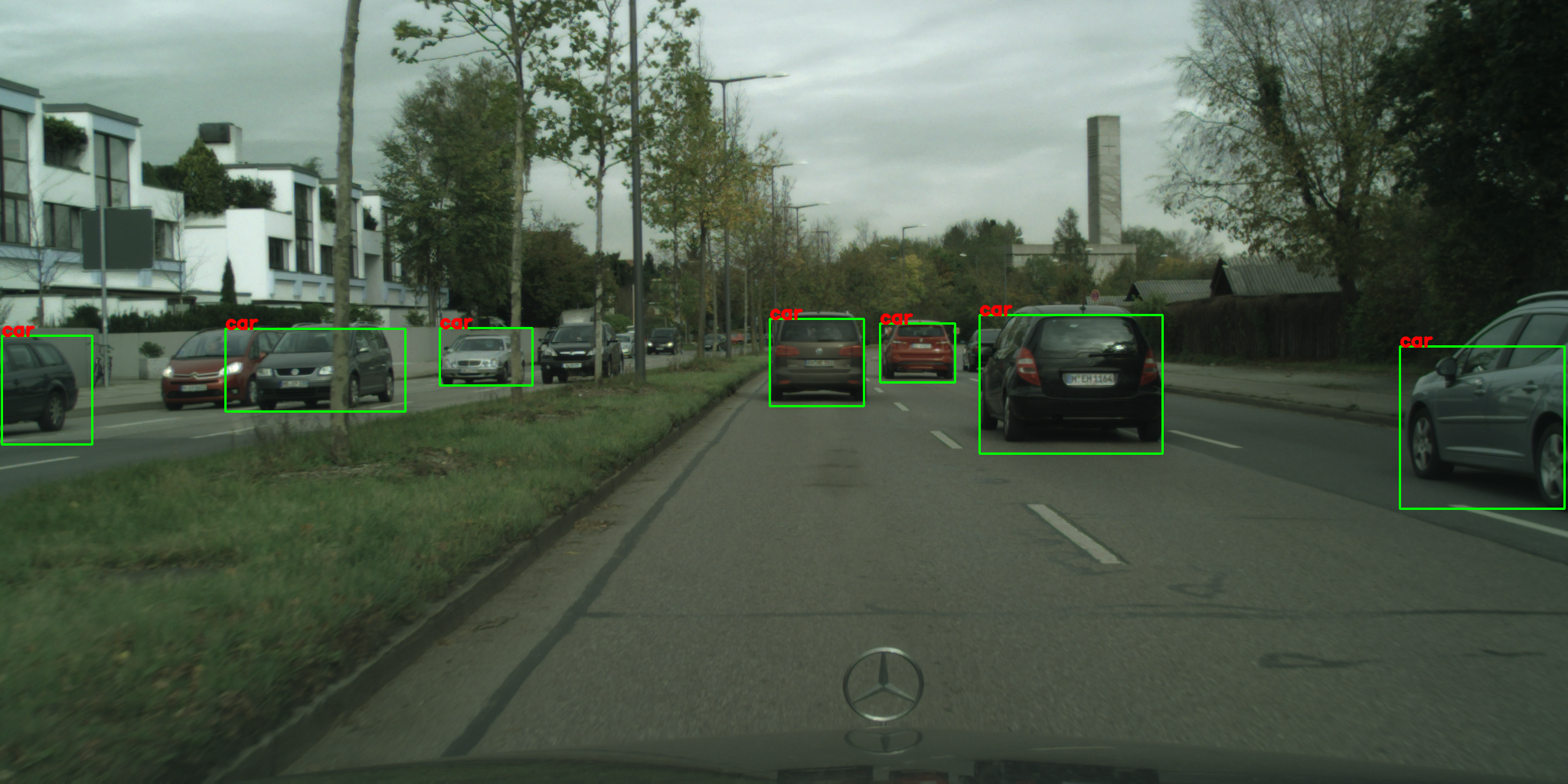}
  \end{subfigure}
  \caption{\small Four continuous output frames of 2DACCAR-1 taking predicted depth images and RGB images from CitySca as inputs}
  \label{fig:image_group_1}
\end{figure}

\clearpage
\newpage
\bibliographystyle{unsrt}

\end{document}